# OntoVerbal: a Generic Tool and Practical Application to SNOMED CT

Shao Fen Liang
Biomedical Research Centre,
NIHR GSTT and King's College London
London, SE1 3QD, UK

Donia Scott
School of Engineering and Informatics,
University of Sussex, Falmer,
Brighton, BN1 9QH, UK

Robert Stevens
School of Computer Science,
The University of Manchester, Oxford Road,
Manchester, M13 9PL, UK

Alan Rector
School of Computer Science,
The University of Manchester, Oxford Road,
Manchester, M13 9PL, UK

*Abstract*—Ontology development is a non-trivial task requiring expertise in the chosen ontological language. We propose a method for making the content of ontologies more transparent by presenting, through the use of natural language generation, naturalistic descriptions of ontology classes as textual paragraphs. The method has been implemented in a proof-of-concept system, OntoVerbal, that automatically generates paragraph-sized textual descriptions of ontological classes expressed in OWL. OntoVerbal has been applied to ontologies that can be loaded into Protégé and been evaluated with SNOMED CT, showing that it provides coherent, well-structured and accurate textual descriptions of ontology classes.

## I. INTRODUCTION

Ontologies and terminologies are increasingly authored in languages based on Description Logics [1] such as the W3C-recommended Web Ontology Language, OWL [2], which support formal descriptions and definitions. This approach has two main benefits. First, the description of entities is explicit within the terminology – for example, what the authors mean by the concept of 'heart disease' can be made explicit and can be interpreted directly by software rather than depending on each user's interpretation of the natural language label *"heart disease"*. Second, where no predefined term exists, new descriptions can be formed by composing expressions using existing classes and properties – *e.g.*, a new class of heart complications caused by emerging diseases (SARS, AIDS, etc.) or new drugs or environmental agents. However, the benefit of using description logics comes at the cost of cognitive complexity and unfamiliar notation. For example, the rendering of the concept of 'heart disease' could be:

'Heart Disease' *EquivalentTo*
('Disorder of Cardiovascular System') *and*
(is-located-in *some* 'Heart Structure')

Ontological descriptions can be much more complex than this, comprising conjunctions of statements that themselves include other nested statements.

While such descriptions are explicit, they can be hard for humans to understand – even those who are trained ontologists.

Partly for this reason, some ontologies are annotated with natural language definitions associated to the logical definitions. These give an alternative view on the main entities of an ontology that avoids potentially impenetrable presentations since are easier to understand, especially when written in the style of natural language is that used by the community in question. For instance, the example above could be annotated with the following text:

*A heart disease is a disorder of the cardiovascular system that is found in a heart structure.*

Writing such natural language definitions by hand is, however, time consuming and there is no guarantee that the meanings of the natural language definitions are the same as the formal logical definitions in the description logic [3]. Also, writing definitions by hand only works for predefined ('pre-coordinated') definitions. When definitions arise from compositional use of pre-coordinated concepts (i.e., 'post-coordinated' definitions), there is no opportunity to write them by hand. Additionally, ontology authoring is heavily reliant on the intervention of 'expert' ontologists, who themselves require many months of experience and training.

One potential solution to this problem is to use existing technologies, such as Natural Language Generation (NLG) from computational linguistics, to produce these natural language descriptions automatically.

This paper describes OntoVerbal, a generic tool that generates automatically natural language descriptions of ontological definitions in ontologies written in OWL. Our aim is to develop a system to produce coherent, reasonably fluent natural language versions of a class' axiomatisation. The potential advantages of OntoVerbal for ontology users are:

*Familiarity:* People may be able to access the content of the ontology without having to learn the (often complex)

This work is part of the Semantic Web Authoring Tool (SWAT) project (see www.swatproject.org), which is supported by the UK Engineering and Physical Sciences Research Council (EPSRC) grant EP/G032459/1, to the University of Manchester, the University of Sussex, and the Open University.

ontology-language in which it is written (e.g., various dialects of OWL, RDF* etc.);

*Browsing:* Navigating an ontology could be facilitated with a browser that provides the option for natural language presentations of selected parts;

*Checking*: Consistency- and error-checking are tedious and difficult tasks, relying on deep expertise in the domain that is being modelled and the ontology-language in use. Having access to natural language descriptions of selected parts of an ontology could make it easier to identify the peculiarities and errors in an ontology;

*Flexible views:* Ontologies are consulted for a range of purposes and by a range of users. Natural language generation provides unique opportunities for tailored presentations of a given ontology (or part thereof), whether by changing the focus of the content or indeed the specific natural language in which it is viewed;

*Training:* Taken together, the above facilities can provide a rich environment for training in ontology writing.

A key design choice in developing the OntoVerbal system was whether to develop (a) a specialised tool that would produce the best possible natural language for a specific ontology or (b) a generic tool that would produce useful but not necessarily perfect natural language for any OWL-EL ontology (the OWL profile that is more widely used in biomedical terminologies), or at least any ontology for a given field of interest. We chose to be generic, and thus to limit OntoVerbal to what is contained in the ontology itself, with minimum recourse to external linguistic resources or tailoring for specific ontologies. Inevitably, this choice means that the language generated is sometimes stilted and unnatural. OntoVerbal thus produces *acceptable* but not always *perfect* English for any OWL-EL ontology.

Similarly, in the representation of OWL's semantics, while we aim to capture the intuitive meaning of the axioms, it is not our intention to provide a 'tutorial' in OWL semantics. So, unlike ACEView [4], which will, for example, spell out the full implications of a transitive property, OntoVerbal does not verbalise all of OWL's semantics to 'explain' the axioms in its input.

Finally, we aim to produce verbalisations of the descriptions of individual classes rather than summarise an entire ontology. Ontology summarisation is an altogether different task [5], and in any case, it is hard to know what it would mean to 'summarise' an entire ontology of the size of many biomedical ontologies – tens or even hundreds of thousands of classes and millions of axioms.

Our chosen approach thus represents three sets of trade-offs: (a) generic applicability with a minimum of additional resources *vs* greater polish of the generated text, (b) comprehensibility of the generated text *vs* complete representation of the OWL semantics, and (c) class-by-class verbalisation *vs* any attempt to summarise the ontology as a whole.

## II. BACKGROUND

Several initiatives in the field of natural language generation (NLG) have addressed the problem of generating better, more coherent and easily processable texts derived from ontologies — for example, ILEX [6], M-PIRO [7], NaturalOWL [8] and Rabbit [9]. These systems make use of annotated data or users' interactions to construct sentences and paragraphs. However, using annotated data and user interaction for text generation does not necessarily help the overall task of ontology comprehension: manually annotating axioms with information to guide language generation works well, but it is time-consuming and requires skill and training; reliance on user input for tasks such as sentence-ordering presents similar issues. Our approach does not preclude the use of such resources to improve the language if so desired, but it does not require it.

Ontologies define the entities in a domain of interest. They are typically authored and presented in groups of axioms (known as 'frames,' sometimes referred to as 'concepts') relating to a single entity. Strictly speaking, the order of axioms in OWL is irrelevant to its meaning, and there is no formal notion of a frame. However, most OWL editors (e.g., Protégé [10], TopBraid Composer [11], Swoop [12] and the NeOn Toolkit [13]) group axioms together into frames as an organisational device to aid modelling and comprehension.

This suggests that grouping sets of axioms is a useful notion. The intuitive textual correlate of a single axiom is a sentence, and that of a frame is a paragraph. When authoring and reading ontologies in natural language, therefore, we focus on paragraphs: the coherence of a textual description of an ontology class, and consequently its comprehensibility, would be increased by grouping ontology (axiom-) sentences together in to topics or units of thought such as (frame-) paragraphs. Within such paragraphs, avoiding repetition by aggregating sentences that conform to regularities should also help.

On top of this, it would also be helpful to know and exploit the way in which many different axioms are asserted in an ontology, since these suggest what are natural or commonplace schemas in ontology construction and thus may provide useful indications of how the corresponding texts should be structured and ordered.

## III. ONTOLOGY AXIOMS

In order to present textually the content of ontologies in a well-ordered, well-structured and fluent manner, we need a clear understanding of the groupings and orderings of axioms that are most typical of ontologies; in other words, we need to know how the ontology community use ontology axioms as a language for describing a domain.

### A. Classifying axioms

If we read axioms from a natural language point of view, we could find that different axioms play different communicative roles. For example:

---

* http://www.w3.org/TR/2004/REC- rdf- mt- 20040210/

*SubClassOf:* indicates that one class is a sub class of another class; these axioms place classes into a taxonomic structure.

*EquivalentClass*: indicates that the classes listed are equivalent; these axioms provide definitions of classes.

*DisjointClass:* indicates that one class is different from other classes; these axioms describe the distinctiveness between classes.

*ClassAssertion:* indicates that an instance is a member of a class; these axioms provide an illustration of a class.

*DisjointUnion*: indicates that a set of classes are exhaustive and are all distinct from each other; these axioms present the alternatives within a given class.

Given the distinctiveness of axioms' respective communicative function, each category of axiom will obviously require different expressions when they are translated into natural language text. A further, orthogonal, classification to consider is the following:

*Simple axioms:* state relations between named classes. For example,

'Disease' (disorder) SubClassOf Clinical finding (finding)

*Complex axioms:* contain not only named classes, but also properties, cardinalities or value restrictions, or combinations of named classes in anonymous class expressions, such as:

'Heart Disease' *EquivalentClass* ('Disorder of Cardiovascular System') *and* RoleGroup some (Finding site *some* 'Heart Structure')

Simple axioms map quite naturally onto simple, declarative, natural language sentences, while complex axioms are likely to be reflected in sentences that are more syntactically and rhetorically complex, and which are often also longer.

*B. Presenting axioms*

The most common views for ontology editing tools to present each class are usage-based and frame-based views. The usage-based view will present to the user every axiom relating to a designated class, whereas tools that make use of the frame-based view will present axioms in distinct categories (e.g., equivalent classes, super classes, class members or disjoint classes). The usage-based view has the advantage of completeness, but the disadvantage of having an unstructured presentation (since axioms are presented in an unordered and unstructured manner). The frame-based view, on the other hand, while making use of a clear structure and ordering, does not provide complete information about the designated class, and the user is thus left to carefully check the complete set of axioms to achieve a full account of the designated class.

Our chosen approach is to model our natural language generation process using the best of these two worlds: for completeness we will follow the usage-based view, presenting all axioms relating to a given class; for comprehensibility, we will follow the frame-based view, by providing a clear structure to the set of axioms.

*C. Structuring axioms*

Given the diversity, complexity and variety of communicative goals of the axioms in an ontology, the issue of how best to present them as a comprehensible text is not trivial. To guide this process, we have undertaken an extensive survey of how common axiom groups relate to a designated class.

TABLE I. LABELS USED TO CLASSIFY AXIOM GROUP

| Axiom group | Simple | Complex |
|---|---|---|
| ClassAssertion | Ca | Car |
| DisjointClass | Dc | Dcr |
| EquivalentClass | Ec | Ecr |
| SubClassOf | Sc | Scr |

We first developed a simple code for labelling axiom groups, assigning two characters to each simple axiom group and three characters to each complex axiom group (see examples in TABLE I). The two characters of each simple axiom group are the first and second capital letters from the axiom type; so for example:

ClassAssertion(John-Joe, Person) is labelled Ca.

The complex axiom groups have an additional character "r"; so for example:

ClassAssertion(John-Joe, Person (and has-Gender(male))) is labelled Car.

We focussed our survey on a set of 490 ontologies collected from the TONES Repository [*], Swoogle [†], and Ontology Design Patterns[‡]. Their Unified Resource Identifiers (URIs) were checked to ensure that our collection was a non-redundant set. Our task here was to determine how many of the axiom groups in this set belong to a designated class.

We therefore gathered, for each of the above-mentioned ontologies, all axioms that contained a designated class and grouped them according to their assigned categories. So, for example a designated class containing axioms of Ec, Scr and Ecr categories would be labelled as EcEcrScr (ordering the categories alphabetically).

This process led to the identification of more than 60 patterns from 268,969 classes of the 490 ontologies, showing that a class can be represented in a variety of structures ranging from a single simple axiom group to a combination of several simple and complex axiom groups. The most common pattern contains only Sc axioms, and is found in over 56% of the classes. The frequency drops sharply to 16.68% for individual Scr axioms and 13.59% for the combination ScScr. Together, these three patterns account for over 87% of identified classes. This indicates that SubClassOf axioms between named classes are the most commonly used for describing ontology classes. Axioms relating to EquivalentClass (containing Ec or Ecr) form the second most common group, containing the patterns Ec, Ecr, EcrSc, and EcrScScr, but do not appear frequently (at only 1 – 5%).

---

[*] http://owl.cs.manchester.ac.uk/repository/

[†] http://swoogle.umbc.edu/

[‡] http://ontologydesignpatterns.org/ont/

At this stage we are certain that taxonomy description is the axiom type that is mostly used by ontologists for describing classes, and a combination of using definition and taxonomy description comes next. According to this finding we then examine our data according to axiom's communicative roles by counting their frequencies from the collected patterns.

*D. Ordering axioms*

As can be seen from **Error! Reference source not found.**, patterns that involve describing taxonomic structures (i.e., Sc + Scr) account for 97% of the cases, followed by groups involving definitions (Ec + Ecr), which account for 75%. Following this are groups conveying distinctions between classes (Dc + Dcr, 49%), providing illustrations (Ca + Car, 44%), and finally those presenting alternatives (Du, 2%).

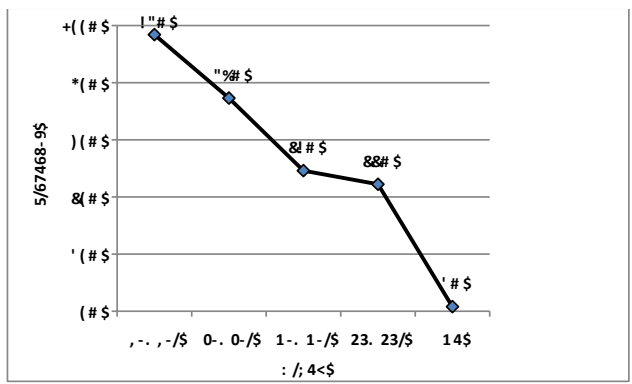

Fig.1.    Order Analysis: communicative role

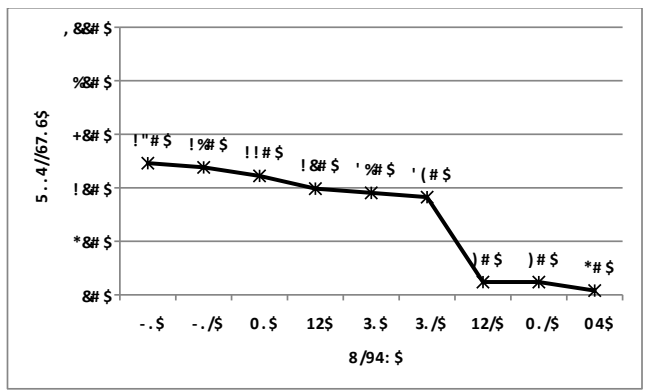

Fig.2.    Order Analysis: axiom complexity

Based on this analysis, we gear our generation algorithms to present axiom groups in the order of their communicative function, as follows: taxonomy, definition, distinctions, illustrations and alternatives.

However, axioms conveying these communicative functions can contain simple or complex axiom types, and attention needs to be paid to this factor during the NLG process. For this, we need to have a principled way of deciding the order in which simple and complex groups should be presented. We therefore undertook a further analysis to separate the axiom groups shown in **Error! Reference source not found.** into their simple and complex groups, and then calculated the occurrence of each group in all patterns. The results, shown in **Error! Reference source not found.**, suggest the following preferred ordering: Sc, Scr, Dc, Ca, Ec, Ecr, Car, Dcr, Du.

As can be seen, although six groups (Sc, Scr, Dc, Ca, Ec, Ecr) occur with some regularity, the remaining (Car, Dcr, Du) hardly ever occur. Since **Error! Reference source not found.** is derived from **Error! Reference source not found.**, we cannot assume that just because Ca occurs more frequently than Ec it should occur before Ec. Rather, we need to focus on each simple group that has a higher occurrence than its corresponding complex group so that, for example, Sc is higher than Scr; Dc is higher than Dcr etc.

Based on these results, we tuned our NLG engine to describe ontology classes by starting with simple axioms before presenting complex ones, and to order axiom groups as follows: Sc, Ec, Dc, Ca, Scr, Ecr[*]. These orderings, based as they are on empirical data on the typical patterns that are used by ontology authors, should be good indicators of what could be 'naturalistic' orderings in the generated paragraphs.

IV. TRANSFORMING AXIOM GROUPS INTO COHERENT TEXT

As with any NLG system, our task begins by organising the input content in such a way as to provide a structure that will lead to coherent text, as opposed to a string of apparently disconnected sentences.

Given the nature of our problem, we need to focus on the semantics of the discourse that can accommodate the nature of ontology axioms. For this purpose, we have chosen to use Rhetorical Structure Theory (RST) [14], as a mechanism for organising the ontological content of the axiom input.

RST is a theory of discourse that addresses issues of semantics, communication and the nature of the coherence of texts, and plays an important role in computational methods for generating natural language texts [15]. According to the theory, a text is coherent when it can be described as a hierarchical structure composed of text spans linked by rhetorical relations that represent the relevance relation that holds between them such as ELABORATION, CONDITION and LIST. Relations can be left implicit in the text, but are more often signalled through discourse markers words or phrases such as "because" for EVIDENCE, "and" for LIST, "or" for ALTERNATIVE, etc. [16; 17]. They can also be signalled by punctuation (e.g., a colon for elaboration, comma between the elements of list, etc.).

---

[*] Given the low occurrence of the groups Car, Dcr, Du and since we do not intend to cover all possible axiom groups at this stage, we have chosen to exclude them for the time being.

Our exploration of RST has shown that some relations appear to map well to the characteristic features of ontology axioms. For example:

- the LIST relation captures those cases where a group of axioms in the ontology bear the same level of relation to a given class;
- the ELABORATION relation applies generally to connect different notions of axioms to a class (i.e., super-, sub- and defining- classes), in order to provide additional descriptive information to the class;
- The CONDITION relation generally applies in cases where an axiom has property restrictions.

Our experience and the evidence over many practical cases have indicated that the full set of rhetorical relations is unlikely to be applied to ontology verbalisation. In particular, the set of so-called presentational relations [18] are unlikely to apply, as ontology authors do not normally create comparisons or attempt to state preferences amongst classes. In addition, even within the set of informational relations, there are several that will not be found in ontologies. For example, since each axiom is assumed to be true, using one axiom as an evidence of another axiom would be redundant. Similarly, using one axiom to justify another axiom is not a conventional way of building ontologies.

*A. From Axiom-sentences to Axiom-paragraphs*

As mentioned earlier, the common approach for translating ontology axioms to natural language is to translate one axiom per sentence [19]. This approach often leads to repetitions and other infelicities; for example, a class can have many sub-classes and translating these sub-class statements into a string of sentences will lead to text that is not only inelegant through its repetition, but also tedious to read [20]. Psycholinguists have long shown that such texts impose an unnecessary cognitive overhead for the reader (see e.g., [21]). Therefore, combining sentences becomes an important issue for avoiding monotony and to aid ease of comprehension. This process is a core linguistic task for any natural language generator that aims to produce fluent text [22], and is thus one that we attempt to utilise in our work.

Transforming individual sentences or clauses into a single, complex sentence involves a process of compounding sentences that focuses on combining subjects, objects and verbs from component sentences or adding punctuation between clauses. In ontology axioms, we can find constructs that map directly onto the linguistic notions of subject, object and verb. For example, the axiom A(X, P) has X as its subject and P as its object; A presents a predicate that holds between X and P, typically expressed in English through a verb. This allows us to apply linguistic operations of sentence (or clause) combining, commonly referred to within computational linguistics as *aggregation* [23; 24], to strings of axioms. Thus, if we have several axioms in the SubClassOf (Sc) group, then we can combine their subjects to generate a compound sentence. Consider, for example, the following three axioms represented as individual sentences:

SubClassOf (X, P) → "X is a kind of P."
SubClassOf (X, Q) → "X is a kind of Q."
SubClassOf (X, R) → "X is a kind of R."

Through the process of aggregation they can be combined to make a single sentence by keeping the same subject and removing the repetition between subject and object, then using a "comma" and an "and" (both discourse markers of the LIST relation that holds between the three axioms) to join objects as the following sentence:

"X is a kind of P, Q and R."

We can extend this approach to produce a range of other types of expressions. For example, if the ontology also included the axiom

SubClassOf (Z, X) → "Z is a kind of X."

that introduces an indirect relation to the subject X, we can use a simple linguistic operation (equivalent to making an active sentence into a passive) to swap subject and object and replace the predicate with its inverse and in doing so produce an appropriate textual expression:

SubClassOf (Z, X) → "A more specialised kind of X is Z."

By iterating the process of aggregation with the complex sentence that we already produced above, we are able to derive a two-sentence paragraph with a consistent focus to cover all four axioms as:

"X is a kind of P, Q and R. A more specialised kind of X is Z."

If we analyse the RST relation in the above example, we will find that its two sentences are in an ELABORATION relation. This is a simple example of the feasibility of transforming axiom-sentences to axiom-paragraphs using RST relations. However, it only illustrates the SubClassOf group. Thus, the next step is to build an RST schema to cover patterns that are required for our purpose.

*B. Building an RST schema for ontology classes*

Our first step in building an RST schema for describing an ontology class is to examine axiom groups in more detail to enable the process of combining multiple axioms into complex sentences (i.e., through aggregation). First of all, we refine the key axiom groups identified at the end of Section D by splitting them into their direct (signalled with a subscript 1) and indirect (signalled with a subscript 2) counterparts — all, that is, except axioms relating to ClassAssertion (Ca), since they are all in direct relations with the designated class (F). The resulting 11 advanced groups are then placed into our RST schema (

):

- the simple-direct category contains SubClassOf (Sc1), EquivalentClass (Ec1) and DisjointClass (Dc1) axioms in their simple and direct forms;

- the complex-direct category contains SubClassOf (Scr1) and EquivalentClass (Ecr1) axioms in their complex and direct forms;
- the simple-indirect category contains SubClassOf (Sc2), EquivalentClass (Ec2) and DisjointClass (Dc2) axioms in their simple and indirect forms;
- the complex-indirect category contains SubClassOf (Scr2) and EquivalentClass (Ecr2) axioms in their complex and indirect forms;
- ClassAssertion (Ca) axioms form a category of their own category.

The axiom groups belonging to each category follow the ordering of the axiom group suggested by the analysis in **Error! Reference source not found.**. As we have illustrated before, SubClassOf axioms belonging to the simple-indirect category ($Sc_2$) can be converted to simple-direct ($Sc_1$) by swapping its subject and object; the same is true for DisjointClasses ($Dc_2$). For example, the axiom:

DisjointClass (A, B, C, F)

is in a direct relation to A, but in an indirect relation to F. If we want to change the focus to F, we can transform the axiom to become:

DisjointClass (F, A, B, C),
DisjointClass (F, B, C, A)

and so on. Such transformations do not affect the underlying meaning of the axiom, but they do change the focus of the resulting natural language expression, so that, for example:

"F is different from A, B and C."

Has the same meaning as

"F is different from B, C and A."

Turning now to our method of expressing axiom classes in English, we use a template-based NLG technique. Our choice in this is driven by an attempt to translate each axiom such that we preserve its meaning in the ontology and avoid introducing misleading information.

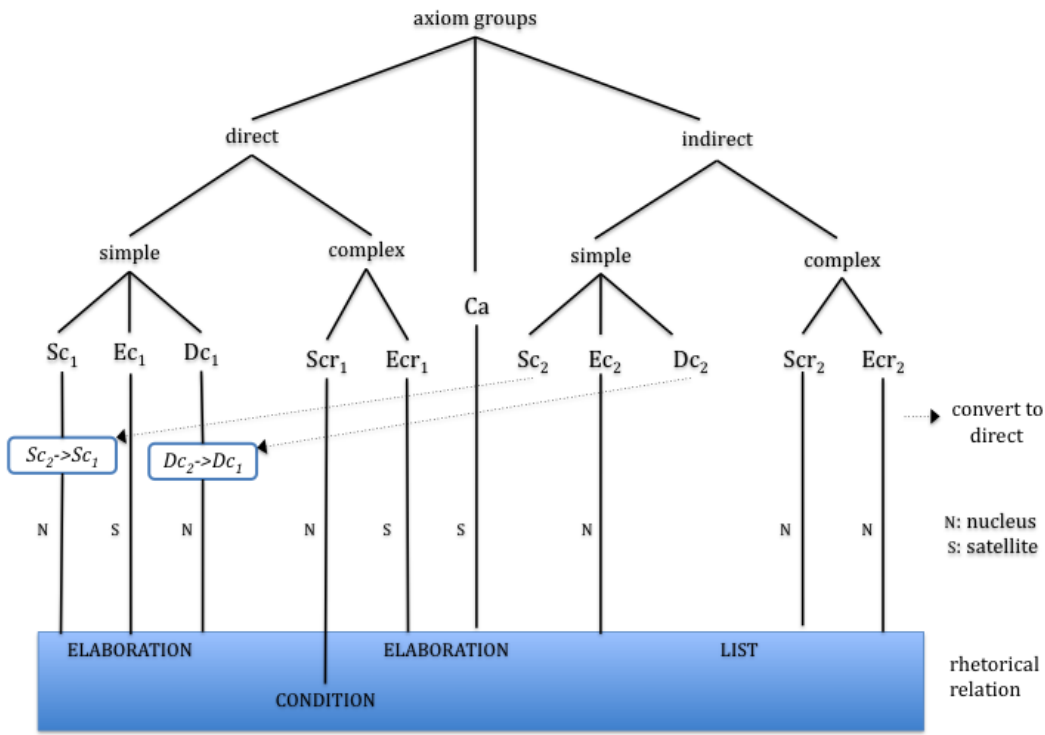

Fig.3. top level RST schema for ontology classes

For example, sub- and super-classes are usually in an 'is a' relation. However, translating them into the English string "is a" can lead to misunderstanding since that expression can be used in English to mean "equal to" or "is the same as"; clearly, though, a class is not equal to or the same as its super-class. In this context, a more accurate translation is "is a kind of" or "is a type of".

*C. The RST paragraph*

In the simple-direct category, $Sc_1$ and the converted $Sc_2$ groups are always the starting description of a class and form the main Nucleus of the paragraph within a top-level elaboration relation. $Ec_1$ is the associated satellite, followed by $Dc_1$ and the converted $Dc_2$. When verbalised, the resulting text is along the lines of:

"F is a kind of X and Y. A more specialised kind of F is Z, and F is defined as P and Q. Also F is different from R."

In the complex-direct category, $Scr_1$ is the Nucleus and the $Ecr_1$ is its Satellite within a condition relation. As the Ca group is on its own, it has been combined into this category as a satellite within an ELABORATION relation. Together, this would lead to a verbalisation such as:

"F is a kind of L that..., and is defined as M that..., and has members $N_1$, $N_2$ ... $N_m$."

We use "additionally" to connect these two categories to make sentences coherent, which leads to a verbalisation such as:

"F is a kind of X and Y.... Additionally, F is a kind of L that . . . $N_m$."

The last part is the simple-indirect and complex-indirect categories. All groups in this part are nuclei, and are in a LIST relation since each is an independent description. As these groups are in an indirect relation to the designated class, the subject in this part of the text is no longer the same as in the previous texts. Thus, we need to introduce some connection words to improve the coherency. For example:

"Another relevant aspect of F is . . . " (for singular)

"Other relevant aspects of F are: . . . " (for plural)

These groups may contain several indirect and complex axioms and without a clear boundary between these axioms, the text can be difficult to read. However, our use of RST, together with the Theory of Discourse Structure [25], allows us to introduce further discourse markers, and even layout elements such as bullet points to make the logical structure of the axiom class more apparent. So the overall content of a class could look like:

"F is a kind of X and Y.... Additionally, F is a kind of L that . . . $N_m$. Other relevant aspects of F are:
• T is a kind of U that ... something to do with F...;
• V is defined as W that ... something to do with F...;
…."

## V. ONTOVERBAL

We have implemented the above methods in a proof-of-concept system, OntoVerbal that takes as input OWL ontologies and produces as output a translation from the OWL into English[*].

As mentioned earlier, OntoVerbal is a generic tool, and thus the output is not 'pretty' English, in that it can be stilted, pedantic and repetitive – characteristics that often come from the nature of the input itself.

For example, ontologies are inherently repetitive and often overly explicit compared to what we would expect of 'good' text (an ontology is likely to state, for example, that a primary school is a type of school or that a left hand is a type of hand); if we remain true to our goal of achieving fidelity *vis a vis* the input, the generated text will necessarily contain these stylistic infelicities.

Another limiting factor is that, while clearly English, the names of ontological concepts are often technical telegraphese; for example, the names of concepts in the SNOMED CT ontology [26; 27] tend not to trip lightly off the tongue – names such as "*bone structure of clavicle and/or scapula and/or humerus*", or "*hypertensive heart and renal disease complicating and/or reason for care during the puerperium*".

*A. OntoVerbal Paragraphs*

We have tested OntoVerbal with a range of ontologies including:

a) The Travel ontology[†], which describes travelling modes, destinations, the boundaries of the destinations and so on.

b) A module of SNOMED CT[‡], which describes medical terminology covering most areas of clinical information such as diseases, findings, procedures, microorganisms, substances, etc.

c) The Experimental Factor Ontology (EFO)[§], which describes experimental variables (e.g. disease state, anatomy) based on an analysis of such variables used in the ArrayExpress database.

These examples of OntoVerbal's input and output are shown in TABLE II. Although our textual paragraphs are not in perfect English, these ontology-entity-embedded-paragraphs are familiar to ontologists.

While our method allows for further improvement of the fluency of the generated paragraphs; the extent to which such improvements are desirable or necessary depends on individual needs. Nevertheless any improvements that do not require heavy additional resources would be welcome. In what follows we describe how we have achieved this, with reference to SNOMED CT.

*B. Tuning OntoVerbal for SNOMED CT*

SNOMED CT is a terminology used for coding in health records and is mandated for use for various purposes in numerous countries around the world[**].

Although the native form of SNOMED CT is a description logic that predates OWL, it is available as OWL and conforms to the OWL EL profile (with the exclusion of disjointness). SNOMED CT provides a useful test-bed for OntoVerbal: it is axiomatically rich, it has no natural language definitions, it has a large user base, and it is employed at many stages by users who have limited experience in OWL.

---

[*] In fact, the system is multilingual, producing both English and Mandarin, but we will focus only on the English here.

[†] http://swatproject.org/ontologies.asp

[‡] For this we used the tool at http://owl.cs.manchester.ac.uk/snomed/ with the signature 'hypertension' .

[§] www.ebi.ac.uk/efo/

[**] http://www.ihtsdo.org

TABLE II. EXAMPLES OF ONTOVERBAL'S INPUT AND OUTPUT

| Travel ontology | |
|---|---|
| Input | (Settlement SubClassOf AdministrativeDivision) (City SubClassOf Settlement) (Town SubClassOf Settlement) (Village SubClassOf Settlement) |
| Output | A settlement is a kind of administrative division. More specialised kinds of settlement are city, town and village. |
| **SNOMED CT ontology** | |
| Input | (Benign hypertensive renal disease (disorder) SubClassOf Hypertensive renal disease (disorder)) (Benign arteriolar nephrosclerosis (disorder) SubClassOf Benign hypertensive renal disease (disorder)) (Benign hypertensive heart AND renal disease(disorder) SubClassOf Benign hypertensive renal disease (disorder)) (Benign hypertensive renal disease (disorder) SubClassOf (Hypertensive renal disease (disorder) and (RoleGroup some (Finding site (attribute) some Kidney structure (body structure))))) |
| Output | Benign hypertensive renal disease(disorder) is a kind of hypertensive renal disease (disorder). More specialised kinds of benign hypertensive renal disease(disorder) are benign arteriolar nephrosclerosis (disorder) and benign hypertensive heart and renal disease (disorder). Additionally, benign hypertensive renal disease (disorder) is a kind of hypertensive renal disease (disorder) that rolegroup a finding site (attribute) a kidney structure (body structure). |
| **Experimental Factor Ontology** | |
| Input | (caudate nucleus SubClassOf cranial ganglion) (caudate nucleus SubClassOf (part of some basal ganglion)) |
| Output | A caudate nucleus is a kind of cranial ganglion. Additionally, a caudate nucleus is a kind of part of a basal ganglion. |

There are many versions of SNOMED CT. We tested OntoVerbal with the July 31 2010 version of the International SNOMED CT. Since the complete ontology is large (292012 classes, 62 object properties) we focussed on a module on hypertension (high blood pressure[*]). Hypertension has many causes and effects, so this module contains a wide range of diseases and anatomic structures: it comprises 506 concepts, each corresponding to a separate OWL class[†]. In addition to its unique identifier which is a 64 bit integer, each concept/class has two associated names: a 'fully specified name' and a 'preferred term' which are natural language expressions such as those shown in TABLE III.

TABLE III. Examples of SNOMED CT names

| SNOMED CT ID no. | Fully-specified name | Preferred name |
|---|---|---|
| 118698009 | Procedure on abdomen (procedure) | Procedure on abdomen |
| 280129003 | Disorder of soft tissue of thoracic cavity (disorder) | Disorder of soft tissue of thoracic cavity |
| 63337009 | Lower trunk structure (body structure) | Lower trunk structure |
| 11511004 | Hypertensive heart AND renal disease complicating AND/OR reason for care during puerperium (disorder) | Hypertensive heart AND renal disease complicating AND/OR reason for care during the puerperium |

For simplicity, we make use of the preferred name. However, the naming conventions for SNOMED CT are complex — not least because this terminology is the result of combining two earlier terminologies, and many different editors and authors have been involved — and this poses particular challenges and limitations for the verbalisation process. As can be seen in TABLE III, names are often expressed in a kind of 'telegraphese', often involving missing articles; they can also be quite long and logically complex. Since concepts/classes obviously map onto nouns in natural language, when names such as these are used verbatim as part of the verbalisation process, they can lead to rather awkward text.

To achieve fully fluent verbalisation would require decoding names into their underlying semantics and re-generating them as more contextually appropriate nominal expressions; however, this would require making use of both linguistic and domain knowledge that is not available in the ontology itself.

For example, the process would have to model (a) the complex mapping between nouns and their articles (i.e., when to use "the", "a", "an") and (b) human anatomy. Armed with the information that the body contains only one heart and one pelvis but several branches of the aorta and several arteries, the verbaliser would then be able to produce the appropriate expressions: "the heart", "the pelvis", but "a branch of the abdominal aorta" and "an artery", etc. Given that our aim is to achieve a generic verbalisation tool, this is clearly beyond the scope of our work — although were such a translation module to exist, OntoVerbal could make use of it[‡]. In the absence of such a module, we have created a somewhat crude version by (a) relying on the use of articles as applied to anatomical terms in Wikipedia[§] and (b) consulting the official list of text definitions for SNOMED CT[**] to find, for cases where the name of an anatomical entity in the 'preferred' name does not include an article, another instance of the same entity which does include one (e.g., "central nervous system" in one instance, but "the central nervous system" in another). When any missing articles were found, they were added to the SNOMED CT input. With the help of these adjustments, we are able to reduce to some extent the awkwardness of the verbalised concept names. In all other respects, though, the verbalisation of the SNOMED CT ontology makes use of the same natural language generation engine that applied to the other ontologies that we have tested.

## VI. EVALUATION

OntoVerbal has now been implemented as a Protégé plugin[††] that can offer an alternative textual view of a class

---

[*] For this we used the tool at http://owl.cs.manchester.ac.uk/snomed/ with the signature 'hypertension'.

[†] The OWL version includes a construct – RoleGroup – that is relevant to only a small number of concepts but has to be included in all for consistency. The Hypertension module was chosen, in part, because RoleGroups were always irrelevant, and they were ignored.

[‡] Such a process has been undertaken to great effect in the Spanish version of SNOMED CT.

[§] http://en.wikipedia.org/wiki/List_of_human_anatomical_features

[**] Given in the file named sct1 TextDefinitions en-US 20110731.txt from the SNOMED CT Technical Implementation Guide that can be downloaded via http://ihtsdo.org/fileadmin/user upload/doc/directory.html

[††] OntoVerbal is available to downloaded from http://swatproject.org/demos.asp

alongside the Manchester syntax view and the graphical views provided by various plugins. We feel the need to undertake a formal evaluation to address two key issues:

**Fidelity:** Are the generated paragraphs faithful to their input? In other words, does the textual output of OntoVerbal convey a clear and true expression of that which is contained in the corresponding input? One way to test this is through a 'round-trip' study: given only the generated output, can a proficient ontologist re-create the semantics of the input from which it was derived?

**Quality:** Are they of reasonable quality? Ideally, the generated paragraphs should be of a standard that is not far from that which one would expect from a proficient ontologist given the task of rendering the input as text.

*A. Experimental set up*

We addressed these questions through an on-line experiment with OntoVerbal applied to SNOMED CT. To avoid making the study too easy or too hard by including very short or very long paragraphs, we narrowed the set under consideration to typical classes in the SNOMED CT ontology, which we found to contain between 3 and 5 axioms. From these we randomly selected 10 for the study.

We created two textual versions of each of the 10 selected classes (see Appendix A): one version was created by OntoVerbal; the other was written by an independent expert ontologist proficient in OWL (with Manchester syntax) [*], under instructions to "transform them into 'fluent' English paragraphs that (a) are semantically equivalent to the OWL and (b) another OWL expert could in principle use to re-create the original OWL". The ontologists were instructed to use the SNOMED CT labels verbatim (through cut-and-paste); this means that if there was a missing article in the SNOMED CT labels (e.g., "artery of the abdomen" rather than "an artery …" or "the artery …"), they would reproduce this in the text – as would OntoVerbal.

We used these 20 texts to design two sets of materials: Set A contained verbalisations of classes 1–5 by the ontologist and 6–10 by OntoVerbal; Set B contained the verbalisations in the other order. With these materials, we conducted an on-line experiment, collecting data from 30 participants who were fluent in OWL EL with Manchester syntax. Half of the participants received Set A, and the others Set B. Each was shown all 10 verbalisations in random order (per participant), with instructions to write the equivalent OWL code. They were instructed to use the SNOMED labels (which were highlighted in the text) verbatim rather than attempt to transform them, and were allowed to cut-and-paste them directly into their code.

*B. Results*

We analysed participants' responses by comparing the code they produced to the OWL input that led to each (machine and human) verbalisation. Since there will be a number of semantic equivalents to each SNOMED CT class descriptions[†], for each of the 10 chosen class descriptions we created all their semantic equivalents. For each response to the presented paragraphs, we measured the Levenshtein distance [28] between the code produced by participants and each member of the set of semantically equivalent versions of the OWL input to that paragraph[‡]. To normalise for the length length of class descriptions, we applied the following operation:

Similarity = (Length of string − Levenshtein distance) / Length of string

Which returns a value of 1 if the code produced is a perfect match to the input (or one of its semantic equivalents), and 0 if there is no match at all. Since the order of axioms in a class description is not a relevant factor, we treated each axiom independently, and registered the mean value over all axioms in the class. Finally, for each response by each subject, we recorded only the highest score received against all members of the set of semantically-equivalent versions of the OWL input. The results of this analysis are shown in TABLE IV.

*1) Fidelity of OntoVerbal's output*

If the paragraphs produced by OntoVerbal are a clear and true expression of the OWL code that it receives as input, participants should be able to re-create the input code or some semantically equivalent version of it. This will be reflected in Similarity scores that are close to 1. Our results show that the mean score over the 10 class descriptions is 0.94. This is an extremely encouraging result, indicating that participants in the study were able to successfully 'translate' the paragraphs generated by OntoVerbal back into the OWL from which they were derived. From this we can conclude with confidence that the output of OntoVerbal is faithful to its input, in the sense that it conveys the correct semantics, since a 'round-trip' is clearly achievable.

TABLE IV. MEAN SIMILARITY SCORES FOR THE 10 CLASS DESCRIPTIONS

| Class | OntoVerbal | Ontologist |
|-------|------------|------------|
| 1     | 0.94       | 0.69       |
| 2     | 0.93       | 0.80       |
| 3     | 0.92       | 0.88       |
| 4     | 0.95       | 0.84       |
| 5     | 0.95       | 0.78       |
| 6     | 0.94       | 0.86       |
| 7     | 0.89       | 0.81       |
| 8     | 0.93       | 0.95       |
| 9     | 0.96       | 0.83       |
| 10    | 0.94       | 0.85       |
| Mean  | 0.94       | 0.83       |

---

[*] The main author of the EFO ontology (www.ebi.ac.uk/efo).

[†] For example, the axiom "A SubClassOf B and C and D" would have 16 semantically equivalent versions.

[‡] We ignored all differences relating to layout (e.g., line breaks or indentation) or to case and punctuation (e.g., 'SubClassOf:' vs 'subclass of').

*2) Quality of the verbalizations*

The high Similarity scores achieved for the output of OntoVerbal suggest that the texts it produces are of good quality for the purpose for which they are intended. Another strong test of the quality of the verbalisations produced by OntoVerbal is the extent to which they compare with verbalisations produced by the expert ontologist, given the same task; one would hope that the mean score for the two versions of each class description (i.e., machine vs human author) would be close. Our results show that although participants were able to translate successfully the paragraphs written by the ontologist (mean Similarity score is 0.83), their performance was consistently below that for the generated texts. Comparing the two statistically, the difference (human *versus* OntoVerbal) is highly significant (mean diff = -.106, t (two-tailed)= -8.025, p<.001).

In other words, the machine-generated verbalisations were better suited for a 'round-trip' than the (probably 'better English') human-written equivalents. As mentioned, the materials for the study were divided into sets, so that all participants saw all 10 paragraphs, but no participant saw both the human-written and OntoVerbal-generated versions of any given paragraph.

This was intended to reduce any bias arising from the style or naturalness of the texts presented, and indeed our statistical analysis of this show no significant effect of the set (mean diff = -.033, t (two-tailed)= -2.055, p<.06).

## VII. Conclusion

The question we sought to address in this work is whether ontology classes can be transformed into readable and reasonably fluent natural language text by an automatic process that is itself reasonably generic. Our experience has provided positive proof that this is the case. We have shown that natural language generation (NLG) technology, enhanced by a discourse planner based on Rhetorical Structure Theory, can transform classes represented in OWL into coherent and fairly fluent paragraphs even in the face of strong constraints (e.g., retaining the textually awkward labels in the ontology and eschewing the use of special purpose linguistic resources). Our NLG architecture does not, of course, require these constraints to be in place: as more sophisticated processes and resources are added, the resulting texts will become closer to that normally found in everyday use. However, the question of just how 'natural' the generated text should be remains open. Indeed, our evaluation study has shown that in some contexts at least, 'natural' (as in human-generated) is not always best. One suspects that the preferred style of the generated text will depend on its intended use, for example, whether for ontology -checking, -browsing, -authoring or training. An advantage of the architecture we have developed for the NLG task of translating OWL classes into text is that it provides the flexibility to tune the style of the generated text.

We have tested our approach in a proof-of-concept system, OntoVerbal, which has also been applied to several ontologies that cover a number of domains, demonstrating that coherent verbalisations can be produced for ontologies within the portion of OWL roughly corresponding to OWL EL. Elsewhere we have shown that the approach also works for other natural languages [29].

Most presentations of OWL ontologies take the form of an OWL syntax along with a visualisation of the ontology's graph or a (manually-written) textual summarisation of the ontology. Verbalisations of the classes in an ontology, such as those provided by OntoVerbal, offers another style of presentation and one that could be a useful counterpart to the overview afforded by a typical graphical presentation. Visualisation tools such as OWLViz[*] and OntoGraf[†] show the the classes in an ontology, but do not give much detail: OWLViz shows only the subclass hierarchy, and OntoGraf does not discriminate between the different quantifications on properties. A class verbalisation could fit neatly into this range of presentations: it gives a presentation of detail, but in a form familiar to users. It is possible to imagine a hybrid presentation with graphical overviews, textual summaries, and textual presentations of the classes' axiomatisation. Protégé, for example, supports many graphical visualisers (described in [30]) that are used to support a number of tasks.

Finally, although our effort has focussed on OWL, there is no a priori reason why it could not be extended to an arbitrary Resource Description Framework (RDF) graph: OntoVerbal is topic-centric (grouping axioms around a given topic) and RDF graphs have the same mechanism (grouping axioms on common URI), thereby making it possible to extract a graph on a topic and verbalise it. Mapping the RST roles onto triples in an RDF graph outside RDFS is, however, an open question.

References

[1] F. Baader, I. Horrocks, and U. Sattler. "Description logics as ontology languages for the semantic web". Lecture Notes in Artificial Intelligence, vol. 2605, pp. 228–248, 2005.

[2] I. Horrocks, P.F. Patel-Schneider, and F.v. Harmelen. "From SHIQ and RDF to OWL: The making of a web ontology language". Journal of Web Semantics, vol. 1, pp. 7-26, 2003.

[3] R. Stevens, J. Malone, S. Williams, R. Power, and A. Third. "Automating generation of textual class definitions from OWL to English". Journal of Biomedical Semantics, vol. 2(Suppl 2):S5, 2011.

[4] T. Kuhn. "The understandability of owl statements in controlled english". Semantic Web journal, 2012.

[5] E. Motta, P. Mulholland, S. Peroni, M. d'Aquin, J.M. Gomez-Perez, V. Mendez, and F. Zablith. "A Novel Approach to Visualizing and Navigating Ontologies". Proceedings of In International Semantic Web Conference, ISWC 2011, Springer-Verlag, 2011.

[6] O.D. Michael, C. Mellish, J. Oberlander, and A. Knott. "ILEX: an architecture for a dynamic hypertext generation system". Natural Language Engineering, vol. 7, pp. 225-250, 2001.

[7] A. Isard, J. Oberlander, C. Matheson, and I. Androutsopoulos. "Speaking the users' languages". IEEE Intelligent Systems Magazine, vol. 18, pp. 40-45, 2003.

[8] D. Galanis, G. Karakatsiotis, G. Lampouras, and I. Androutsopoulos. "An open-source natural language generator for OWL ontologies and its use in Protégé and second life". Proceedings of 12th Conference of

---

[*] http://www.co-ode.org/downloads/owlviz/

[†] http://protegewiki.stanford.edu/wiki/OntoGraf

the European Chapter of the Association for Computational Linguistics (EACL'09), pp. 17-20, 2009.

[9] G. Hart, M. Johnson, and C. Dolbear. "Rabbit: developing a Control Natural Language for authoring ontologies". Proceedings of 5th Annual European Semantic Web Conference (ESWC 2008), pp. 348–360, 2008.

[10] D.L. Rubin, N.F. Noy, and M. Musen. "Protégé: a tool for managing and using terminology in radiology applications". Journal of Digital Imaging,, vol. 20, pp. 34-46, 2007.

[11] M. Erdman. "Ontology engineering and plug-in development with the NeOn Toolkit". Proceedings of 5th Annual European Semantic Web Conference (ESWC 2008), 2008.

[12] A. Kalyanpur, B. Parsia, E. Sirin, B.C. Grau, and J. Hendler. "Swoop: a web ontology editing browser". Journal of Web Semantics, vol. 2, pp. 144–153, 2006.

[13] D. Allemang, and I. Polikoff. "TopBraid, a multi-user environment for distributed authoring of ontologies". Proceedings of 3rd International Semantic Web Conference (ISWC 2004), Springer Verlag 2004.

[14] W.C. Mann, and S.A. Thompson. "Rhetorical Structure Theory: toward a functional theory of text organisation". Text, vol. 8, pp. 243-281, 1988.

[15] C. Mellish, A. Knott, J. Oberlander, and M. O'Donnell. "Experiments using stochastic search for text planning". Proceedings of 9th International Workshop on Natural Language Generation, pp. 98-107, 1998.

[16] C.B. Callaway. "Integrating discourse markers into a pipelined natural language generation architecture". Proceedings of 41st Annual Meeting on Association for Computational Linguistics, pp. 264-271, 2003.

[17] C. Sporleder, and A. Lascarides. "Using automatically labelled examples to classify rhetorical relations: an assessment". Natural Language Engineering, vol. 14, pp. 369-416, 2008.

[18] M. Moser, and J.D. Moore. "Toward a synthesis of two accounts of discourse structure". Computational Linguistics, vol. 22, pp. 409–420, 1996.

[19] A. Third, S. Williams, and R. Power. "Owl to english: a tool for generating organised easily-navigated hypertexts from ontologies". Proceedings of 10th International Semantic Web Conference (ISWC 2011), 2011.

[20] S. Williams, and R. Power. "Grouping axioms for more coherent ontology descriptions". Proceedings of 6th International Natural Language Generation Conference (INLG 2010), pp. 197–201, 2010.

[21] H.H. Clark. "Psycholinguistics". MIT Press. 1999.

[22] C. Mellish, D. Scott, L.C.D Paiva, R. Evans, and M. Reape. "A reference architecture for natural language generation systems". Natural Language Engineering, vol. 12, pp. 1–34, 2006.

[23] H. Dalianis. "Aggregation as a subtask of text and sentence planning". Proceedings of Florida AI Research Symposium, FLAIRS-

[24] M. Reape, and C. Mellish. "Just what is aggregation, anyway?". Proceedings of European Workshop on Natural Language Generation, 1999.

[25] R. Power, D. Scott, and N. Bouanyad-Agha. "Document structure". Computational Linguistics, vol. 29, pp. 211-260, 2003.

[26] K.A. Spackman, and K.E. Campbell. "Compositional concept representation using SNOMED: Towards further convergence of clinical terminologies". Journal of the American Medical Informatics Association, pp. 740-744, 1998.

[27] M.Q. Stearns, C. Price, K.A. Spackman, and A.Y. Wang. "SNOMED clinical terms: overview of the development process and project status". Proceedings of AMIA Fall Symposium (AMIA-2001), Henley & Belfus, pp. 662-666, 2001.

[28] V.I. Levenshtein. "Binary codes capable of correcting deletions, insertions and reversals. ". Sov. Phys. Dokl., vol. 6, pp. 707-710, 1966.

[29] S.F. Liang, R. Stevens, and A. Rector. " OntoVerbal-M: a Multilingual Verbaliser for SNOMED CT". Proceedings of 2nd International Workshop on the Multilingual Semantic Web (MSW 2011) in conjunction with the International Semantic Web Conference (ISWC2011), pp. 13–24, 2011.

[30] A. Katifori, C. Halatsis, G. Lepouras, C. Vassilakis, and E. Giannopoulou. "Ontology Visualization Methods—A Survey". ACM Computing Surveys, vol. 39, 2007.

C. *Appendix A:* Verbalisations Of OWL Input By OntoVerbal And By An Expert Ontologist

| Number | OWL input | OntoVerbal | Ontologist |
|---|---|---|---|
| 1 | pelvic structure SubClassOf: lower trunk structure<br><br>lower trunk structure SubClassOf: structure of subregion of the trunk<br><br>pelvic structure SubClassOf: the pelvis and lower extremities and the abdomen and the pelvis and lower trunk structure | A lower trunk structure is a kind of structure of subregion of the trunk. A more specialised kind of lower trunk structure is pelvic structure. Another relevant aspect of lower trunk structure is that a pelvic structure is defined as the pelvis and lower extremities, the abdomen and the pelvis and a lower trunk structure. | A lower trunk structure is a structure of the subregion of the trunk. A pelvic structure is a subtype of a lower trunk structure. |
| 2 | procedure on the brain SubClassOf: intracranial procedure<br><br>intracranial procedure SubClassOf: procedure on the central nervous system<br><br>intracranial procedure SubClassOf: procedure on the head<br><br>intracranial procedure EquivalentClass procedure by site and has a procedure site some intracranial structure | An intracranial procedure is a kind of procedure on the central nervous system and procedure on the head. A more specialised kind of intracranial procedure is procedure on the brain. Additionally, an intracranial procedure is defined as a procedure by site that has a procedure site in an intracranial structure. | An intracranial procedure is a procedure on the central nervous system and a procedure on the head. Any procedure by site in which the procedure site is an intracranial structure is also an intracranial procedure. A procedure on the brain is a subtype of intracranial procedure. |
| 3 | abdominal vascular structure SubClassOf: abdominal and pelvic vascular structure<br><br>abdominal and pelvic vascular structure SubClassOf: vascular structure of the trunk<br><br>abdominal vascular structure SubClassOf: abdominal structure and abdominal and pelvic vascular structure | An abdominal and pelvic vascular structure is a kind of vascular structure of the trunk. A more specialised kind of abdominal and pelvic vascular structure is abdominal vascular structure. Another relevant aspect of abdominal and pelvic vascular structure is that an abdominal vascular structure is defined as an abdominal structure and an abdominal and pelvic vascular structure. | An abdominal and pelvic vascular structure is a vascular structure of the trunk. An abdominal vascular structure is a subtype of an abdominal and pelvic vascular structure. |
| 4 | chronic disease of the genitourinary system SubClassOf: | Chronic disease of the genitourinary system | A chronic disease of the |
| | chronic disease<br><br>chronic disease of the genitourinary system SubClassOf: disorder of the genitourinary system<br><br>chronic hypertensive uraemia SubClassOf: chronic disease of the genitourinary system<br><br>chronic disease of the genitourinary system EquivalentClass: chronic disease and disorder of the genitourinary system and has a finding site some structure of the genitourinary system | is a kind of chronic disease and disorder of the genitourinary system. A more specialised kind of chronic disease of the genitourinary system is chronic hypertensive uraemia. Additionally, chronic disease of the genitourinary system is defined as chronic disease that is a disorder of the genitourinary system, and has a finding site in a structure of the genitourinary system. | genitourinary system is a chronic disease and a disorder of the genitourinary system. Any chronic disease which is also a disorder of the genitourinary system and is found in the structure of the genitourinary system is also a chronic disease of the genitourinary system. A chronic hypertensive uraemia is a subtype of a chronic disease of the genitourinary system. |
| 5 | finding of the head and the neck region SubClassOf: finding of the body region<br><br>head finding SubClassOf: finding of the head and the neck region<br><br>finding of the head and the neck region EquivalentClass: finding of the body region and has a finding site some head and neck structure<br><br>head finding EquivalentClass: finding of the head and the neck region and has a finding site some head structure | A finding of the head and the neck region is a kind of finding of the body region. A more specialised kind of finding of the head and the neck region is head finding. Additionally, A finding of the head and the neck region is defined as a finding of the body region that has a finding site in a head and neck structure. Another relevant aspect of finding of the head and the neck region is that a head finding is defined as a finding of the head and the neck region that has a finding site in a head structure. | A finding of the head and the neck region is a finding of the body region. Any finding of the body which is found in a head and neck structure is also a finding of the head and neck region. A head finding is a subtype of the finding of the head and the neck region. |
| 6 | nephrosclerosis SubClassOf: degenerative disorder<br><br>degenerative disorder SubClassOf: disease<br><br>arteriosclerotic vascular disease SubClassOf: degenerative disorder<br><br>degenerative disorder EquivalentClass: disease and has an associated morphology some degenerative abnormality | Degenerative disorder is a kind of disease. More specialised kinds of degenerative disorder are nephrosclerosis and arteriosclerotic vascular disease. Additionally, degenerative disorder is defined as disease that has an associated morphology in a degenerative abnormality. | A degenerative disorder is a disease. Any disease which has an associated morphology of degenerative abnormality is also a degenerative disease. Nephrosclerosis and arteriosclerotic vascular disease are subtypes of degenerative disease. |
| 7 | kidney graft material SubClassOf: urinary tract material<br><br>kidney graft material SubClassOf: solid organ graft material<br><br>kidney graft material SubClassOf: urinary tract material and solid organ graft material<br><br>transplant of the kidney EquivalentClass: kidney operation and solid organ transplant and renal replacement and has a method some surgical transplantation action and has a direct substance some kidney graft material and has an indirect procedure site some kidney structure | A kidney graft material is a kind of urinary tract material and solid organ graft material. Another relevant aspect of kidney graft material is that a transplant of the kidney is defined as a kidney operation that is a solid organ transplant, and is a renal replacement, and has a method in a surgical transplantation action, and has a direct substance in a kidney graft material, and has an indirect procedure site in a kidney structure. | Kidney graft material is a urinary tract material and a solid organ graft material. A kidney operation, solid organ transplant and renal replacement which has a method of surgical transplantation action, a direct substance of kidney graft material and an indirect procedure site of kidney structure is a type of transplant of the kidney. |
| 8 | graft SubClassOf: biological surgical material<br><br>tissue graft material SubClassOf: graft<br><br>tissue graft material SubClassOf: graft and body tissue surgical material | A graft is a kind of biological surgical material. A more specialised kind of graft is tissue graft material. Another relevant aspect of graft is that a tissue graft material is defined as a graft and a body tissue surgical material. | A graft is a biological surgical material. Tissue graft material is a subtype of graft as well as a body tissue surgical material. |
| 9 | benign essential hypertension complicating and/or reason for care during pregnancy SubClassOf: essential hypertension complicating and/or reason for care during pregnancy<br><br>essential hypertension complicating and/or reason for care during pregnancy SubClassOf: essential hypertension in the obstetric context<br><br>essential hypertension complicating and/or reason for care during pregnancy SubClassOf: pre-existing hypertension in the obstetric context<br><br>essential hypertension complicating and/or reason for care during pregnancy SubClassOf: essential hypertension in the obstetric context and pre-existing hypertension in the obstetric context<br><br>benign essential hypertension complicating and/or reason for care during pregnancy SubClassOf: benign essential hypertension in the obstetric context and essential | Essential hypertension complicating and/or reason for care during pregnancy is a kind of essential hypertension in the obstetric context and pre-existing hypertension in the obstetric context. A more specialised kind of essential hypertension complicating and/or reason for care during pregnancy is benign essential hypertension complicating and/or reason for care during pregnancy. Another relevant aspect of essential hypertension complicating and/or reason for care during pregnancy is that benign essential hypertension complicating and/or reason for care during pregnancy is defined as benign essential hypertension in the obstetric context and essential hypertension complicating and/or reason for care during pregnancy. | An essential hypertension complicating and/or reason for care during pregnancy is an essential hypertension in the obstetric context and a pre-existing hypertension in the obstetric context. A benign essential hypertension complicating and/or reason for care during pregnancy is a subtype of essential hypertension complicating and/or reason for during pregnancy. |

| | hypertension complicating and/or reason for care during pregnancy | | |
|---|---|---|---|
| 10 | procedure on artery of the abdomen SubClassOf: procedure on the abdomen<br><br>procedure on artery of the abdomen SubClassOf: procedure on artery of the thorax and the abdomen<br><br>abdominal artery implantation SubClassOf: procedure on artery of the abdomen<br><br>procedure on artery of the abdomen EquivalentClass: procedure on artery and has a procedure site some structure of artery of the abdomen | A procedure on artery of the abdomen is a kind of procedure on the abdomen and procedure on artery of the thorax and the abdomen. A more specialised kind of procedure on artery of the abdomen is abdominal artery implantation. Additionally, a procedure on artery of the abdomen is defined as a procedure on artery that has a procedure site in a structure of artery of the abdomen. | A procedure on artery of the abdomen is a procedure of the abdomen and a procedure on artery of the thorax and the abdomen. Any procedure on artery which has a procedure site of structure of artery of the abdomen is also a procedure on artery of the abdomen. An abdominal artery implantation is a subtype of procedure on artery of the abdomen. |